\documentclass[sigconf,screen,nonacm]{acmart}
\acmConference[Conference acronym 'XX]{Make sure to enter the correct
  conference title from your rights confirmation emai}{June 03--05,
  2018}{Woodstock, NY}

\usepackage{graphics} 
\usepackage{epsfig} 

\usepackage{float}
\usepackage{ragged2e}
\usepackage{caption}
\usepackage{multirow}

\begin{document}

\title{Exercise Specialists’ Evaluation of Robot-led Physical Therapy for People with Parkinson’s Disease}

\author{Matthew Lamsey}
\affiliation{%
    \institution{Georgia Institute of Technology}
    \country{}
}
\email{lamsey@gatech.edu}

\author{Meredith D. Wells}
\affiliation{%
    \institution{Superfeet Worldwide}
    \country{}
}
\email{mwells@superfeet.com}

\author{Lydia Hamby}
\affiliation{%
    \institution{Emory University}
    \country{}
}
\email{lydia.hamby@emory.edu}

\author{Paige Scanlon}
\affiliation{%
    \institution{Emory University}
    \country{}
}
\email{paige.scanlon@emory.edu}

\author{Rouida Siddiqui}
\affiliation{%
    \institution{Emory University}
    \country{}
}
\email{rouida.siddiqui@emory.edu}

\author{You Liang Tan}
\affiliation{%
    \institution{University of California, Berkeley}
    \country{}
}
\email{youliang@berkeley.edu}

\author{Jerry Feldman}
\affiliation{%
    \institution{The Parkinson's Foundation}
    \country{}
}
\email{jhfeldman1@gmail.com}

\author{Charles C. Kemp}
\affiliation{%
    \institution{Hello Robot, Inc.}
    \country{}
}
\email{ck@hello-robot.com}

\author{Madeleine E. Hackney}
\affiliation{%
    \institution{Emory University}
    \country{}
}
\email{mehackn@emory.edu}





\begin{abstract}

Robot-led physical therapy (PT) offers a promising avenue to enhance the care provided by clinical exercise specialists (ES) and physical and occupational therapists to improve patients' adherence to prescribed exercises outside of a clinic, such as at home. Collaborative efforts among roboticists, ES, physical and occupational therapists, and patients are essential for developing interactive, personalized exercise systems that meet each stakeholder's needs. We conducted a user study in which 11 ES evaluated a novel robot-led PT system for people with Parkinson's disease (PD), introduced in \cite{lamsey2023stretch}, focusing on the system's perceived efficacy and acceptance. Utilizing a mixed-methods approach, including technology acceptance questionnaires, task load questionnaires, and semi-structured interviews, we gathered comprehensive insights into ES perspectives and experiences after interacting with the system. Findings reveal a broadly positive reception, which highlights the system's capacity to augment traditional PT for PD, enhance patient engagement, and ensure consistent exercise support. We also identified two key areas for improvement: incorporating more human-like feedback systems and increasing the robot's ease of use. This research emphasizes the value of incorporating robotic aids into PT for PD, offering insights that can guide the development of more effective and user-friendly rehabilitation technologies.

\end{abstract}

\maketitle




\section{INTRODUCTION}

Due in part to a globally aging population \cite{world2015world}, Parkinson’s disease (PD) is now among the fastest-growing neurodegenerative disorders in the United States \cite{feigin2021burden}. As of 2020, over one million Americans are living with PD \cite{yang2020current}. People with PD (PWP) often experience motor impairments such as bradykinesia, hypometria, and postural instability \cite{factor2014freezing, pal2016global, stegemoller2014timed} as well as cognitive impairments and difficulty with dual-tasking \cite{pretzer2009parkinson, litvan2011mds, geurtsen2014parkinson} (i.e., conducting simultaneous mental and motor tasks). Physical therapists (PTs) and occupational therapists (OTs) often treat these symptoms with progressive and tailored physical and cognitive exercises \cite{o2014clinical, salgado2013evidence, schenkman2012exercise} which have been shown to mitigate symptom progression and improve individuals' quality of life \cite{caglar2005effects, ferrazzoli2016does, van2013effects, mak2017long}. However, there are factors, including PWP's difficulties with intrinsic motivation \cite{pickering2013self} and a general shortage of healthcare providers \cite{buerhaus2008current}, that remain obstacles for achieving maximal benefits from correct exercise dosing \cite{clarke2016physiotherapy}.

Many types of exercise have been evaluated as treatments for PD. A review by Mak et. al. discusses several types of exercise used as PD treatments, including gait training, walking exercise, balance exercise, tai-chi, dance, and "exergaming"  \cite{mak2019exercise}. These exercises target specific PD symptoms including hypometria, freezing of gait, coordination, balance, and muscle strength. Diverse exercise interventions spanning dance, aquatic, and cueing training have been shown to improve freezing of gait in PWP \cite{gilat2021systematic}. Some symptoms require more specific interventions, however, gait training has been established as an effective method to reduce fall risk in PWP \cite{protas2005gait}, and PWP's performance of activities of daily living (ADLs) also showed the greatest improvement when training was specifically targeted at ADLs \cite{foster2021occupational}. These findings highlight the importance of personalized exercise interventions and innovative approaches to effectively deliver and optimize therapy for PD. Yet, the required exercise dose for effective results often surpasses what a therapist can provide, even in small groups \cite{clarke2016physiotherapy}, which suggests that supplementing human-supervised exercise with technology-based interventions may enhance motivation and adherence.

\begin{figure*}
    \centering
    \includegraphics[width=\textwidth]{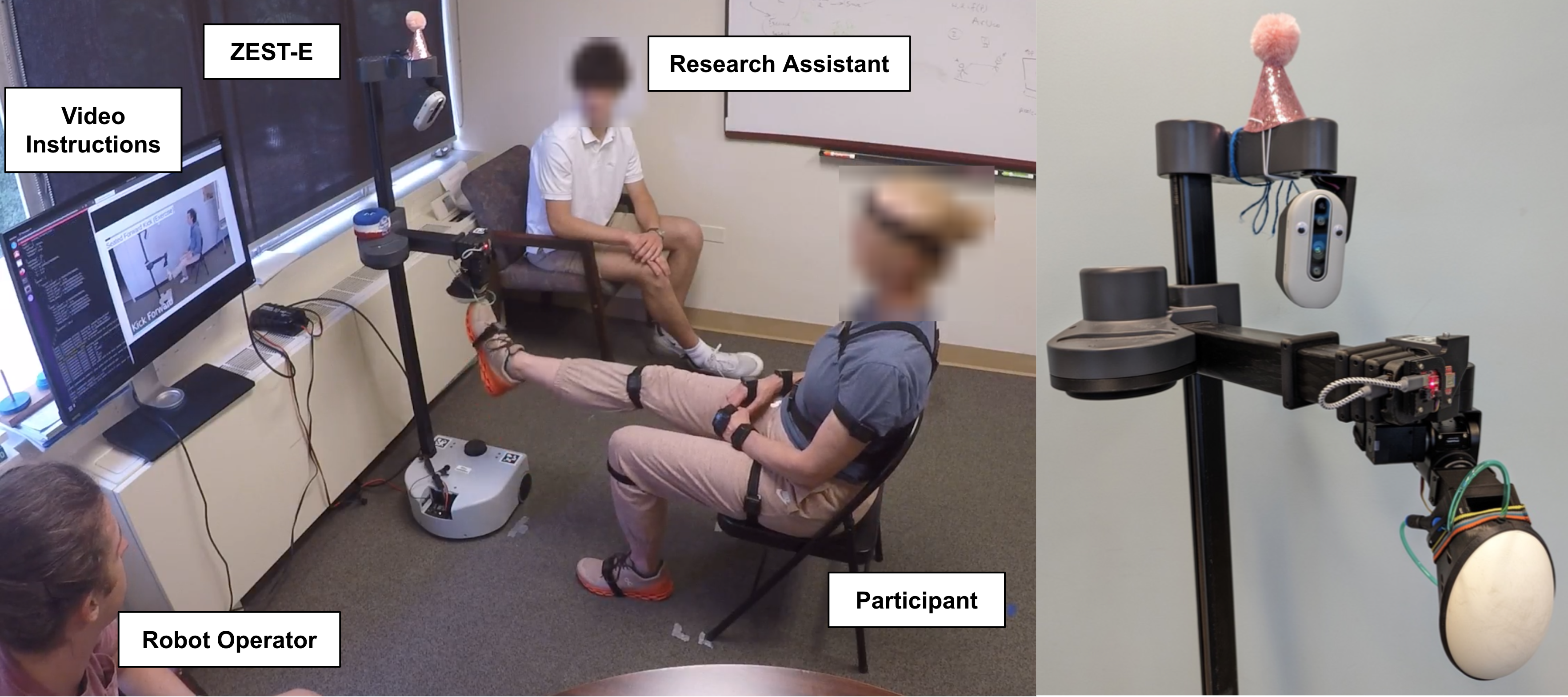}
    \caption{(Left) Experimental setup with an Exercise Specialist (ES) performing a seated forward kick exercise with the Zesty Exercise System for Therapeutic Engagement (ZEST-E). (Right) ZEST-E, a robotic physical therapy system with a soft bubble end effector that serves as an external cue for stretching exercises. }
    \label{fig:sws}
\end{figure*}

Robotic exercise aids offer a promising solution to enhance therapeutic exercises facilitated by PTs and OTs. A 2024 review of robotics in physical rehabilitation \cite{banyai2024robotics} highlights the promise of robotic technologies such as exoskeletons, assistive training devices, and brain-computer interfaces for rehabilitation. Yet, this review cites consistent areas for improvement related to robots' ease of use and high system cost. Physically interactive robotic therapy systems have been tested in interventions for specific impairments, such as post-stroke rehabilitation of gross motor skills \cite{johnson2003design, krebs2004rehabilitation} and fine motor skills \cite{vakili2023impact, urrutia2023spasticity}, walking rehabilitation \cite{wuversatile, regmi2020design}, and motor-cognitive rehabilitation \cite{aprile2020robotic}. These systems combine physical and visual feedback to guide therapeutic exercises, which has proven to be beneficial in alleviating users' symptoms. However, they frequently incorporate purpose-built hardware that is too large, expensive, or complex for use outside of a clinical environment. Wearable rehabilitation robots, such as neck rehabilitation robots \cite{doss2023comprehensive} and hand exoskeletons \cite{tran2021hand}, are more portable, yet remain expensive and purpose-built for a single task.

Socially Assistive Robots (SARs) have also been studied as an often lower cost and more accessible form of robotic therapy, including seated exercise \cite{fasola2012using} and tabletop rehabilitation games \cite{feingold2021robot}. Social-physical human-robot interaction (HRI) aims to combine the benefits of socially and physically interactive systems through exercises such as hand-clapping exercise games \cite{fitter2020exercising} and emotional support via hugging \cite{onishi2024moffuly}. While socially and physically interactive systems are rated as highly engaging, they may present safety concerns due to the robots' large sizes and heavy masses. Augmented Reality (AR) and Virtual Reality (VR) have also been used to supplement therapeutic exercise for PD. While head-mounted AR devices may improve exercise adherence \cite{tunur2020augmented}, other work found that AR has limited utility in treating freezing of gait, and at times can worsen these symptoms \cite{janssen2020effects}. Head-mounted VR systems \cite{campo2022wearable, sanchez2020impact} have been used in limited capacities for gait and balance training for PWP and may result in short-term improvements in motor performance \cite{kwon2023systematic}.

Understanding how different individuals experience a new technology is crucial for developing effective solutions. Tools like Technology Acceptance Models (TAMs) \cite{venkatesh2000theoretical, venkatesh2008technology} capture users’ attitudes, including perceived usefulness and ease of use. The NASA Task Load Index (TLX) \cite{hart1988development, hart2006nasa} assesses mental and physical demands associated with a task, which can be applied to assistive technologies. Similarly, the Perceived Impact of Assistive Devices Scale (PIADS) \cite{jutai2002psychosocial} measures how an assistive device influences competence, adaptability, and self-esteem. Recent mixed-methods studies involving clinicians considering the administration of robotic therapy highlight critical factors for rehabilitation robots, such as telepresence, ease of use, reliability, and cost \cite{sobrepera2022therapists}, and also report generally positive perceptions of these robots’ usefulness and ease of use in stroke rehabilitation \cite{klaic2024application}.

Previous work \cite{lamsey2023stretch} presented a personalized, socially and physically interactive robotic physical therapy system for PWP, originally named “Stretch with Stretch” (SWS) and now referred to as ZEST-E (Zesty Exercise System for Therapeutic Engagement). PWP found that ZEST-E was engaging, and they perceived robot-led exercises to be moderately difficult, which aligns with recommendations for exercise-based interventions for PD \cite{salgado2013evidence}.

The goal of this study was to build on this prior work by evaluating ZEST-E through a mixed-methods study with Exercise Specialists (ES), which we defined as individuals who have professionally administered exercises to PWP. This definition includes physical and occupational therapists, exercise physiologists, and exercise instructors. ES evaluated the system both from the perspective of a potential user and through the lens of potentially prescribing ZEST-E as a treatment for their patients. We hypothesized that ES would provide high ratings for the system based on technology acceptance surveys, and we sought to identify key factors that facilitate or hinder adoption and improvement of systems like ZEST-E.

\section{METHODS}

\subsection{User Study}

Eleven ES, all of whom worked with PWP, participated in an exploratory trial to collect data regarding healthcare providers’ perceptions of ZEST-E. Participants had experience working with PWP in a variety of capacities, including physical therapy, occupational therapy, and strength training. This study was approved by Institutional Review Boards at the Georgia Institute of Technology (protocol H22359) and the Emory University School of Medicine (protocol STUDY00004909). All participants provided written informed consent prior to participation.

\begin{table}[]
    \centering
    \caption{ZEST-E Evaluative Questionnaire items, where $i$ indicates the order in which each item appeared in the survey. Each item was scored on a 5-point Likert scale, where 1 corresponds to Strongly Disagree and 5 corresponds to Strongly Agree. All themes are from the patient perspective, except for Healthcare Provider Considerations.}
    
    \begin{tabular}{|p{0.875in} p{0.1in} p{2.0in}|} \hline
        \textbf{Theme} & $i$ & Questionnaire Entry \\ \hline

        \multirow[c]{6}{=}{\textbf{Usability and Interaction}} & 2 & The robot was easy to play with. \\
        & 4 & The robot clearly communicated what it wanted me to do. \\ 
        & 6 & The robot clearly communicated how I should perform each exercise. \\ 
        & 18 & The robot did what I expected it to do. \\ \hline

        \multirow[c]{5}{=}{\textbf{Engagement and Motivation}} & 3 & The robot provided games that were engaging. \\ 
        & 12 & The robot motivated me. \\ 
        & 13 & The robot encouraged me. \\ 
        & 15 & The robot was engaging. \\ \hline

        \multirow[c]{6}{=}{\textbf{User Experience}} & 11 & The robot was friendly. \\ 
        & 14 & The robot was frustrating. \\ 
        & 16 & The robot was clear. \\ 
        & 17 & The robot was "old school." \\ 
        & 23 & I felt comfortable exercising with the robot. \\ \hline

        \multirow[c]{6}{=}{\textbf{Performance and Accuracy}} & 1 & The robot judged my performance accurately. \\
        & 5 & The robot understood the rate at which I completed the task. \\
        & 7 & The robot provided useful feedback on my performance. \\ \hline

        \multirow[c]{9}{=}{\textbf{Therapeutic Efficacy and Relevance}} & 8 & The robot encouraged me to initiate motions. \\ 
        & 9 & The robot encouraged me to exercise with the proper magnitude. \\ 
        & 10 & The robot encouraged me to exercise with the proper speed. \\ 
        & 21 & The robot was appropriately challenging. \\ 
        & 22 & The robot provided exercise targets at an appropriate distance from me. \\ \hline
        
        \multirow[c]{3}{=}{\textbf{Operational / Environmental Considerations}} & 19 &  The robot did not rush me. \\ 
        & 20 & I had enough space to exercise. \newline \\ \hline

        \multirow[c]{10}{=}{\textbf{Healthcare Provider Considerations}} & 24 & The robot would be useful for my clinic. \\
        & 25 & The robot would offload tasks from me. \\ 
        & 26 & The robot would be useful for my patients to use. \\ 
        & 27 & The robot would be easy for my patients to use. \\ 
        & 28 & The robot would be difficult for my patients to use. \\ 
        & 29 & The robot would provide challenging exercises for my patients. \\ \hline
       
    \end{tabular}
    \label{tab:ZEQ}
\end{table}

Each ES participated in one session lasting for approximately 2.5 hours, following the same protocol as participants with PD \cite{lamsey2023stretch}. The experimental setup for this study is shown in Figure \ref{fig:sws}. ZEST-E led each participant through an exercise session that lasted approximately one hour and consisted of four sets of six stretching exercises, totaling 24 sets. Two of the exercises involved a simultaneous cognitive task. For each exercise, ZEST-E first provided verbal and video instructions for completing the exercise. Then, ZEST-E guided users through a calibration of their right and left side ranges of motion in order to gather data regarding where to initially place the target for each exercise. Lastly, participants performed two 30-second sets of repetitions of the exercise with the right and left side for a total of four sets. ZEST-E adjusted the difficulty of the exercise by moving the location of the target between each set based on the user's performance. After the exercise session, we administered several surveys, and at the conclusion of the session, each participant was led through a semi-structured exit interview about their experience with ZEST-E.

\subsection{Surveys}

Surveys administered after the exercise session included the Technology Attitudes Questionnaire \cite{rosen2013media}, NASA TLX \cite{hart1988development, hart2006nasa}, the Psychosocial Impact of Assistive Devices (PIADS) Scale \cite{jutai2002psychosocial}, and the Robot Opinions Questionnaire \cite{chen2017older}. For the NASA TLX, we computed the "Raw TLX" scores for each of the six task load metrics \cite{hart2006nasa}. For PIADS, we aggregated the responses into three categories rated on a scale of -3 to +3, where -3 corresponds to "strongly decreases" and +3 corresponds to "strongly increases." Each questionnaire was scored according to the methodology in their corresponding original publications. We also administered a custom survey, the ZEST-E Evaluative Questionnaire (Table \ref{tab:ZEQ}), regarding specific features of our system.


We also administered a custom survey, the ZEST-E Evaluative Questionnaire (ZEQ), to evaluate specific attributes of our system. The ZEQ consists of questions from the perspective of both a user of the system as well as questions from that of a clinician who would administer robot-led therapy as part of treatment (Table \ref{tab:ZEQ}). We categorized questions into six distinct themes to capture comprehensive insights into the participants' experiences with the robot. Themes such as "Usability and Interaction," "Engagement and Motivation," and "User Experience" primarily focus on users' perceptions and interactions with the system. Meanwhile, "Performance and Accuracy" along with "Therapeutic Efficacy and Relevance" assess the robot’s effectiveness in leading therapeutic exercises. "Healthcare Provider Considerations" directly assesses how the ES perceive the value of the system in their practice. For each theme in the generic (patient perspective) portion of the ZEQ, we computed Cronbach's $\alpha$ to determine the internal consistency. Highly consistent $\alpha$ values (>0.61 \cite{mchugh2012interrater}) allow us to compute an average response across individual items for each theme for each participant \cite{chen2017older}.

\subsection{Exit Interviews}

In addition to questionnaires, we conducted a semi-structured exit interview with each participant. Questions from the interview are provided in Table \ref{tab:ssi_questions}. The interview was conducted immediately following completion of the surveys. The intent of the interview questions was to determine positive and negative attributes of the system, as well as to indicate areas for improvement and barriers for adoption.

Interviews were coded using NVivo 12 software. First, two independent raters performed open coding to identify overarching themes. A third independent researcher compared the open coding in a group conversation to create the final codebook. The final codebook was refined against one participant's interview until 85\% consistency \cite{chen2017older} and Cohen's kappa $>0.6$ \cite{mchugh2012interrater} were achieved. Two raters then independently coded the remainder of the interviews, and the third researcher provided clarifications during the process. Codes were organized as parent themes with sub-themes, as given in Table \ref{tab:codebook}. Opinions of the ES were often divided for each sub-theme, so we assigned a positive or negative valence to each coded instance of a sub-theme.

\begin{table}[]
    \centering
    \caption{Semi-structured interview questions}
    
    \begin{tabular}{|p{0.15in}  p{2.75in}|} \hline
        1. & How was the overall experience with ZEST-E? \\ \hline
        2. & What did you like about the ZEST-E system?  \\ \hline
        3. & What about the ZEST-E system could be improved?  \\ \hline
        4. & What additional things would you like ZEST-E to do?  \\ \hline
        5. & What additional features would you like ZEST-E to have?  \\ \hline
        6. & What would make you more likely to use the ZEST-E system?  \\ \hline
        7. & For what purpose would you want to use the ZEST-E system?  \\ \hline
        8. & Where would you want to use the ZEST-E system?  \\ \hline
        9. & What type of patient group and/or clientele would you want to use the system (PTs only)?  \\ \hline
        10. & How do you see ZEST-E becoming a system that would be most beneficial to you?  \\ \hline
        11. & How likely are you to recommend this ZEST-E system to someone else?  \\ \hline
        12. & What additional feedback do you have regarding the system or the experience?  \\ \hline
    \end{tabular}
    \label{tab:ssi_questions}
\end{table}

\begin{table}[]
    \centering
    \caption{Codebook for Semi-structured Interviews}
    
    \begin{tabular}{|p{0.875in}|p{0.25in}|p{1.75in}|} \hline
        \textbf{Parent Theme} & ID & Sub-theme \\ \hline
        \textbf{Aesthetics and Design} & AD1 & Robot has an appealing physical design
and accessories \\ \hline
        \multirow[c]{4}{=}{\textbf{Engagement and Motivation}} & EM1 & Robot is engaging, motivating, and fun for PT exercises \\ \cline{2-3}
         & EM2 & Robot could improve exercise adherence at home \\ \hline
        \multirow[c]{4}{=}{\textbf{Quality of Communication}} & QC1 & Robot communicated clearly and effectively \\ \cline{2-3}
         & QC2 & Robot provided useful feedback on users' performance \\ \hline
        \multirow[c]{5}{=}{\textbf{Usability and Reliability}} & UR1 & Robot is portable \\ \cline{2-3}
         & UR2 & Robot is easy for an ES to set up and operate \\ \cline{2-3}
         & UR3 & Robot is easy for PWP to set up and operate \\ \hline
        \multirow[c]{7}{=}{\textbf{Function and Safety}} & FS1 & Robot could adapt to an individual's physical capabilities \\ \cline{2-3}
         & FS2 & Robot could work with PWP with various symptoms \\ \cline{2-3}
         & FS3 & Robot was safe to exercise with \\ \cline{2-3}
         & FS4 & Participants perform robot-led exercise correctly \\ \hline
        \multirow[c]{8}{=}{\textbf{Healthcare Provider Considerations}} & HC1 & Robot seems economically accessible \\ \cline{2-3}
         & HC2 & Robot would improve ES job efficiency \\ \cline{2-3}
         & HC3 & Robot could improve health outcomes \\ \cline{2-3}
         & HC4 & Robot could improve health data collection \\ \hline
        \multirow[c]{9}{=}{\textbf{Expected Use Cases}} & EC1 & Robot would be useful for PWP \\ \cline{2-3}
         & EC2 & Robot would be useful for people with conditions other than PD \\ \cline{2-3}
         & EC3 & Robot would be useful for healthy people \\ \cline{2-3}
         & EC4 & Robot would be useful at home or in a clinic \\ \cline{2-3}
         & EC5 & Robot would be useful somewhere other than a home or a clinic \\ \hline
        \multirow[c]{7}{=}{\textbf{Suggestions and Questions}} & S1 & Amplifying a positive aspect \\ \cline{2-3}
         & S2 & Rectifying a negative aspect \\ \cline{2-3}
         & S3 & Neutral changes to existing features \\ \cline{2-3}
         & S4 & Adding a novel feature \\ \cline{2-3}
         & S5 & Questions and suggestions about a business model \\ \hline
    \end{tabular}
    \label{tab:codebook}
\end{table}

\section{RESULTS}

\subsection{Study Population}

The demographic information for our participant population is given in Table \ref{tab:demographics}.

\begin{table}[]
    \centering
    \caption{Participant Demographics}
    
    \begin{tabular}{|p{0.875in}|p{2.125in}|} \hline
        \textbf{Gender} & 4 Male (36\%), 7 Female (64\%)  \\ \hline
        \textbf{Age} & 23-64 years, $\mu$ = 35, $\sigma$ = 12 \\ \hline
        \multirow[c]{3}{=}{\textbf{Ethnicity}} & 8 White / Caucasian (73\%), 1 Black / African-American (9\%), 1 Asian (9\%), 1 Multiracial (9\%) \\ \hline
        \multirow[c]{3}{=}{\textbf{Education Past High School}} & 1 some college / Associates (9\%), 3 Bachelor’s (27\%), 2 Master’s (18\%), 5 Doctoral (45\%) \\ \hline
        \multirow[c]{2}{=}{\textbf{Occupation}} & 3 exercise instructors (27\%), 7 physical therapists (64\%), 1 exercise physiologist (9\%) \\ \hline
    \end{tabular}
    \label{tab:demographics}
\end{table}

\subsection{Questionnaires}

\begin{figure}
    \centering
    \includegraphics[width=0.45\textwidth]{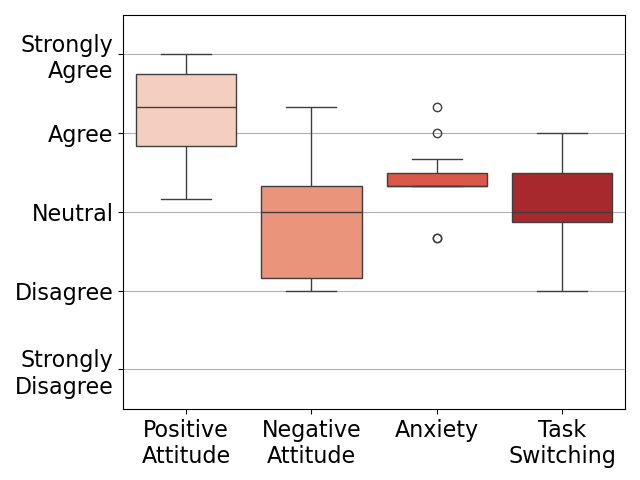}
    \caption{Technology Attitudes Questionnaire \cite{rosen2013media}. Positive attitude: (M = 4.3, IQR = 0.9); Negative attitude: (M = 3.0, IQR = 1.2); Anxiety: (M = 3.3, IQR = 0.2) with outliers \textit{S02} (4.3), \textit{S03} (4.0), \textit{S04} (2.7), and \textit{S05} (2.7); Task Switching: (M = 3.0, IQR = 0.6).}
    \label{fig:tech_attitudes}
\end{figure}

Results from the Technology Attitudes questionnaire are given in Figure \ref{fig:tech_attitudes}, which indicate participants' general attitudes towards technology. Survey responses were aggregated into four themes rated on a Likert-type scale from 1-5, where 1 indicates strong disagreement attitude towards technology and 5 indicates strong agreement with that attitude.

\begin{figure*}
    \centering
    \includegraphics[width=\textwidth]{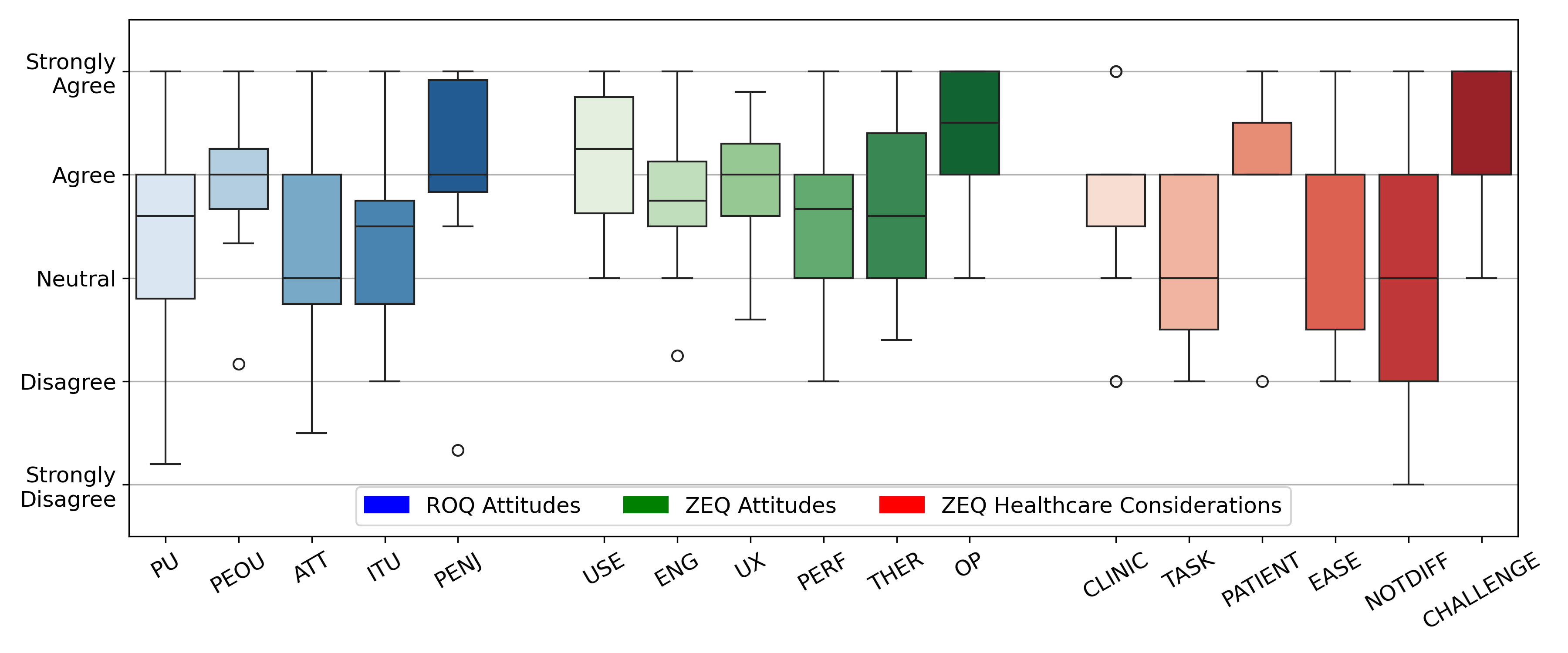}
    \caption{Robot Opinions Questionnaire \cite{chen2017older} (ROQ, Blue) and ZEST-E Evaluative Questionnaire (ZEQ, Green and Red). ROQ: Perceived Usefulness (PU): (M = 3.6, IQR = 1.2); Perceived Ease of Use (PEOU): (M = 4.0, IQR = 0.6) with outlier \textit{S03} (2.2); Positive Attitidue (ATT): (M = 3.0, IQR = 1.25); Intent to Use (ITU): (M = 3.5, IQR = 1.0); Perceived Enjoyment (PENJ): (M = 4.0, IQR = 1.1) with outlier \textit{S08} (1.3). ZEQ Attitudes: Usability and Interaction (USE): (M = 4.25, IQR = 1.1); Engagement and Motivation (ENG): (M = 3.8, IQR = 0.6) with outlier \textit{S08} (2.3); User Experience (UX): (M = 4.0, IQR = 0.7); Performance and Accuracy (PERF): (M = 3.7, IQR = 1.0); Therapeutic Efficacy (THER): (M = 3.6, IQR = 1.4); Operational and Environmental Considerations (OP): (M = 4.5, IQR = 1.0). ZEQ Healthcare Considerations: Usefulness in Clinic (CLINIC): (M = 4.0, IQR = 0.5) with outliers \textit{S01} (2), \textit{S02} (5), \textit{S03} (2), and \textit{S05} (5); Offloading Tasks (TASK): (M = 3.0, IQR = 1.5); Useful for Patients (PATIENT): (M = 4.0, IQR = 0.5) with outlier \textit{S03} (2); Ease of Use for Patients (EASE): (M = 4.0, IQR = 1.25); Difficult for Patients to Use (DIFF): (M = 3.0, IQR = 2.0); Exercise would be Challenging (CHALLENGE): (M = 4.0, IQR = 1.0). Response distributions from the Healthcare Provider Considerations section of the ZEQ are presented individually for each survey item.}
    \label{fig:roq_zeq}
\end{figure*}

Outcomes from the Robot Opinions Questionnaire \cite{chen2017older} (ROQ) and ZEST-E Evaluative Questionnaire (ZEQ) are given in Figure \ref{fig:roq_zeq}. Each of the attitudes was rated on a Likert-type scale from 1-5, where 1 indicates strong disagreement or negative feelings towards an attitude, and 5 indicates strong agreement or positive feelings towards that attitude. Each theme in the ZEQ had a Cronbach's alpha of 0.76-0.89, indicating strong internal consistency.

\begin{figure*}
    \centering
    \includegraphics[width=0.95\textwidth]{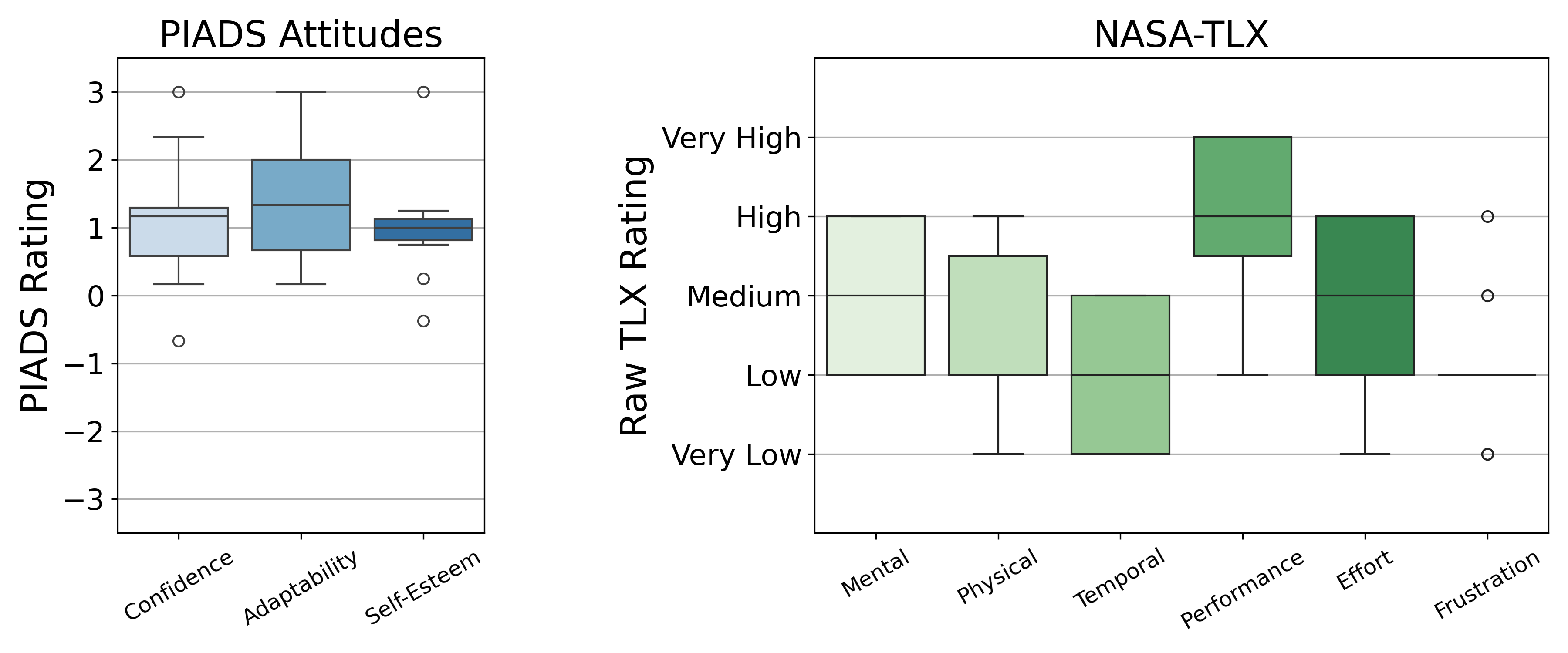}
    \caption{(Left) Psychosocial Impact of Assistive Devices Scale (PIADS) \cite{jutai2002psychosocial}: Confidence: (M = 1.17, IQR = 0.71); Adaptability with outliers \textit{S02} (3.0) and \textit{S03} (-0.7): (M = 1.33; IQR = 1.33); Self-Esteem: (M = 1.0, IQR = 0.31) with outliers \textit{S02} (3.0), \textit{S03} (-0.4), and \textit{S06} (0.3) (Right) NASA TLX survey \cite{hart1988development, hart2006nasa}: Mental: (M = 3, IQR = 2.0); Physical: (M = 2; IQR = 1.5); Temporal: (M = 2, IQR = 2.0); Performance: (M = 4, IQR = 1.5); Effort: (M = 3, IQR = 2.0); Frustration: (M = 2, IQR = 0.0) with outliers \textit{S02} (1), \textit{S06} (3), \textit{S07} (4), and \textit{S08} (1).}
    \label{fig:piads_tlx}
\end{figure*}

Response distributions from the PIADS \cite{jutai2002psychosocial} and the NASA TLX survey \cite{hart1988development, hart2006nasa} are presented in Figure \ref{fig:piads_tlx}.

\subsection{Exit Interviews}

Insights and quotes from the semi-structured exit interviews are arranged by parent theme and sub-theme ID (Table \ref{tab:codebook}). For clarity, quotes coded as suggestions are sorted into their corresponding parent themes. We indicate the number of participants whose responses were coded positively or negatively for each sub theme; for example, sub-theme \textbf{AD1} had two positive and three negative responses, and is notated below as \textbf{(AD1; pos=2, neg=3)}. Participant IDs are specified in parentheses e.g., \textit{(S02)}.



\subsubsection{Aesthetics and Design (AD)} $ $

\textbf{(AD1; pos=2, neg=3)}: Two participants thought the robot’s appearance was unobtrusive, describing how its setup “didn’t feel crowded” in the experiment space \textit{(S06)}. These participants reported that the design was balanced and made the participants feel at ease. Two others expressed concern that the robot "doesn’t scream exercise" \textit{(S05)} or that the robot "might be intimidating for some of my patients" \textit{(S11)}.

\textbf{AD Suggestions:} Improvements to the aesthetics of the system include making the robot's end effector "more approachable" and adding target markings to the end effector (i.e., concentric circles) \textit{(S11)}.

\subsubsection{Engagement and Motivation (EM)} $ $

\textbf{(EM1; pos=8, neg=5)}: Several participants enjoyed the robot's verbal cues to exercise at a specific rate. They also found the robot's spoken encouragements, such as “you’re doing great” and “let’s make it harder,” to be engaging \textit{(S01)}. Others were more apprehensive about whether a robot would be used for exercise at home: "[my patients would] probably just occasionally talk to the robot and then just put it on the side, I think" \textit{(S08)}.

\textbf{(EM2; pos=5, neg=3)}: Exercise specialists were optimistic that the robot could improve their patients' adherence to at-home exercise plans. One participant said that using the robot at home would be their "number one" use case for the technology \textit{(S10)}, and others highlighted the potential for robotic exercise coaches to improve telemedicine for individuals who live far away from clinics \textit{(S04)} or have difficulty leaving their homes \textit{(S01)}. Improving adherence to specific exercise instructions without supervision was a common theme:

\begin{quote}
    This would be great... if they could have one (sic, robot) to go home with them to serve as motivation and also just give specific, clear instructions on how to perform a certain task because a lot of things about [exercise] form are often things that we can't help them with when they're not with us in person, especially with a home-based rehab program. \textit{(S07)}
\end{quote}

In contrast, one participant hypothesized that the robot would be less engaging once its novelty wore off: "I feel like if the system was set in someone's home, they would probably use it one or two times and then it'd start getting [not so] interesting to interact with" \textit{(S08)}.

\textbf{EM Suggestions:} There were several suggestions for improving the engagement and motivating qualities of the system. These include adding accompanying music \textit{(S08)}, increasing the variability of the exercise pacing \textit{(S06)}, and adding more randomness to the motions that the robot directed \textit{(S08)}. One ES proposed adding more auditory and visual cues to the exercises \textit{(S03)}, and another suggested gamification improvements, e.g., adding a high score board, tracking performance between sessions, and adding a visual component to the dual-tasking exercises \textit{(S01)}.

\subsubsection{Quality of Communication (QC)} $ $

\textbf{(QC1; pos=7, neg=3)}: Feedback from ES on the communication quality of ZEST-E was more positive than negative. Positive feedback highlighted that ZEST-E provided "clear, concise instructions" \textit{(S09)} and offered good "encouragement and verbal cues" \textit{(S03)} as well as good "auditory feedback" \textit{(S02)}.

On the other hand, participants who had negative views on the communication quality mentioned that they "had a hard time understanding some of the [robot's] words clearly" \textit{(S06)} and experienced confusion. One participant expressed that they "couldn't figure out what [the robot] wanted from [them]" \textit{(S06)}.

\textbf{(QC2; pos=7, neg=3)}: Participants generally had a positive outlook on the quality of the robot's feedback. They appreciated that the feedback was "interactive," "approachable,"  and "not intimidating" \textit{(S02)}. One participant also found the robot's "positive reinforcement" to be particularly "helpful" \textit{(S01)}. However, some participants felt that the robot's feedback was too pre-programmed \textit{(S10)}.

\textbf{QC Suggestions:} Multiple ES highlighted areas for improving the robot's communication. Regarding the robot's instructions, suggestions included creating a phone app to provide more exercise information \textit{(S07)} and allowing the user to ask the robot to clarify or repeat instructions \textit{(S06)}. Accessibility concerns were raised, such as adding subtitles to the robot's visual instructions, supporting other languages than English, and allowing the user to control the volume of the robot's voice \textit{(S01)}. One participant wished the feedback could be "more dynamic" and "more like a trainer, rather than just passive like, [saying the same thing after every action, regardless of what's going on]" \textit{(S10)}.

Specific improvements to the robot's feedback include automatic detection of incorrect exercise form \textit{(S11)}, more nuanced feedback from the robot after sets beyond exercise score reporting \textit{(S06)}, and general improvements to the "sincerity" of the robot's spoken feedback \textit{(S09)} to make the interactions feel less "forced" \textit{(S10)}. One ES suggested that the robot should specifically encourage users to perform "full motions" during exercise sets \textit{(S11)}.

\subsubsection{Usability and Reliability (UR)} $ $

\textbf{(UR1; pos=2, neg=3)}: Three participants expressed concerns about the portability of the robot. One ES stated that the robot would not be accessible to patients in the home health community due to challenges with "lugging a robot around" \textit{(S01)}. Another highlighted the need to simplify or eliminate the "setup and breakdown" of the system \textit{(S02)}.

\textbf{(UR2; pos=8, neg=4)}: Most participants thought that the system would be easy for ES to set up and operate. They appreciated the combination of visual and verbal cues to teach the exercises. The system was reported to be "very easy to use, very easy to follow" \textit{(S08)}. However, some areas of improvement include further reducing user input during the exercise session \textit{(S03)} and streamlining the robot setup process \textit{(S02)}.

\textbf{(UR3; pos=4, neg=5)}: ES less optimistic about PWP using the system.

\begin{quote}
    I think how comfortable an individual is with technology is a huge component—that cannot be overlooked, especially [since] most Parkinson’s patients are older adults. There's still a very large percentage of older adults that do not use smartphones, that do not use a lot of technology, so they may be very apprehensive if [the robot] is too much. That itself is too overwhelming. It doesn’t matter how great [the robot is]. \textit{(S01)}
\end{quote}

Opinions were divided regarding the role of a caregiver operating the robot. One participant thought that "you could set it up in a way that someone could be pretty independent with using it with relatively low risk of injury" \textit{(S01)}, while another claimed that, "if somebody were to take it home at this stage, they wouldn't be able to operate it independently. And I don't even know that it would be able to be operated with a reasonably educated caregiver either" \textit{(S03)}.

\textbf{UR Suggestions:} Improvements to the usability of the system included making the robot voice-activated or operated with a single button press \textit{(S10)}. One ES expressed a general need for easier independent operation by patients \textit{(S03)}, and another requested the inclusion of "set workouts" that come with the robot \textit{(S05)}. Minimizing setup and breakdown complexity was also emphasized \textit{(S02)}. It was noted that there should be two versions of the system: one for use in a clinic and one for use at home \textit{(S10)}. That particular ES also proposed replacing the robot's external screen with a tablet mounted on the robot \textit{(S10)}.

\subsubsection{Function and Safety (FS)} $ $

\textbf{(FS1; pos=7, neg=6)}: Opinions on the robot's ability to adapt to individual physical capabilities were mixed. Some felt that "pretty much anybody could benefit from those exercises" \textit{(S03)} due to their simplicity. However, others expressed concerns about performing the exercises incorrectly. One participant was unsure whether they should focus on speed or full range of motion, noting, "I felt myself not coming back fully on the chair or not standing up the whole way" \textit{(S11)}. Another participant worried that trying to complete more repetitions might lead to compromised movement quality, such as "not coming all the way back up" \textit{(S01)} during exercises that required large vertical motions.

\textbf{(FS2; pos=3, neg=4)}: There were differing opinions on how well the robot could serve PWP with varying symptoms. Some felt the robot could be effective if it could "calibrate the system or the programs based on the individual needs of the individual at that specific moment in time," considering that PWP might have varying abilities throughout the day and at different "on/off" points in their medication cycles \textit{(S01)}. They also suggested that sessions might need to be "shortened" \textit{(S01)} and adapted for those with severe mobility needs, allowing for breaks and assistance as required. Some participants felt that ZEST-E would be best suited for users who are "relatively cognitively intact" \textit{(S03)} and that it should be designed for "someone at a high enough level to benefit from it, but not so advanced that the exercises would become unnecessary" \textit{(S03)}.

\textbf{(FS3; pos=3, neg=1)}: Participants were split in their views on the robot’s safety. One participant felt that working with ZEST-E "could be pretty independent" with a "relatively low risk of injury" \textit{(S01)}. However, others raised concerns about fall safety. One participant feared that exercises like forward leaning could "throw [patients] off balance" or make them "dizzy," and suggested that additional safety precautions and guidelines might be needed \textit{(S06)}.

\textbf{(FS4; pos=3, neg=4)}: Participants were divided on whether the robot-led exercises were performed correctly. One participant appreciated the auditory feedback, saying it helped users realize they were using their full range of motion, which "you don’t normally get, just like typical exercise" \textit{(S02)}. However, others were concerned about the exercise speed, with one participant unsure if they should prioritize speed or full range of motion \textit{(S11)}. Another participant echoed this worry, expressing concern that focusing on repetitions might compromise the quality of movements, such as not fully returning to the starting position \textit{(S01)}. 

\textbf{FS Suggestions:} Beyond the robot's feedback system, ES had several ideas for expanding the robot's functionality. There was interest in using the robot to help people practice standing up \textit{(S01)} and for gait training or as a walking aid \textit{(S03)}. One ES suggested having the robot monitor fall risks \textit{(S05)}, and multiple ES expressed a general desire for an expanded repertoire of exercises, such as overhead reaching \textit{(S01)}, fine motor skill training \textit{(S01)}, or resistance exercises with a band \textit{(S07)}. Another ES requested a reduction in time between exercise sets to increase intensity \textit{(S10)}. Several ES proposed further personalization options, such as more individualized difficulty and pace \textit{(S04)}, exercises in a larger workspace around the human \textit{(S08)}, calibrating the exercise difficulty based on outcome measures other than range of motion \textit{(S10)}, and automatically detecting the need to take a break \textit{(S01)}.

\subsubsection{Healthcare Provider Considerations (HC)} $ $


\textbf{(HC1; pos=1, neg=3)}: Several participants expressed concern that the system would be prohibitively expensive. Specifically, participants cited a need for robotic treatments to be covered by health insurance to appeal to most users.

\begin{quote}
    Not everybody can afford their medications, so they’re not going to choose this versus pain meds or paying for groceries... It might be exclusionary from that regard. \textit{(S01)}
\end{quote}

\textbf{(HC2; pos=9, neg=3)}: Most participants agreed that ZEST-E could improve their job efficiency inside and outside of the clinic. Helpful applications of a robotic exercise coach that were cited by participants included offloading repetitive tasks \textit{(S10)}, enabling a healthcare provider to work with many patients at once \textit{(S07)}, and serving as an instructional reference for patients to use during home exercise \textit{(S07)}. However, some participants expressed concern that the system would not offload enough work: "if it requires a bunch of input from me as a PT, I'd just rather do [the exercise coaching] myself" \textit{(S03)}.




\textbf{(HC3; pos=5, neg=1)}: Exercise specialists were optimistic that the robot could improve their patients' health outcomes. One participant highlighted the importance of external cues for exercise, stating that "[PWP] think that they’re doing full range of motion with things, but very often they’re not... [the robot has] that auditory feedback that you’re actually using your full range of motion" \textit{(S02)}. Another ES envisioned robot-led exercise as a useful starting point for individuals beginning exercise:

\begin{quote}
    I think, at first, it would be easier to target the more sedentary people [with robot-led exercise] - get them to a level where they are comfortable, so then maybe [they will] seek out other resources, even just getting a gym membership or the personal trainer or things like that. Give them at least the ability to have a good foundation so they’re not starting cold, give them confidence to do that as well. \textit{(S04)}
\end{quote}

\textbf{(HC4; pos=4, neg=2)}: Several individuals highlighted the potential for robots like ZEST-E to augment health data collection. It was noted by ES that the performance measures that the robot collected may need to be improved and expanded to capture more comprehensive aspects of the exercise performance \textit{(S10)}.

\textbf{HC Suggestions:} Providing an interface for healthcare providers to customize the exercise sections \textit{(S11)} and augmenting exercise statistics collected by the robot \textit{(S03)} are two key areas for improvement. One participant noted that, "if it's something that the patient can do without me, that's great, but I don't feel comfortable billing for that" \textit{(S06)}. In general, ES "would have to know the cost" \textit{(S09)} of the system; when one participant was told the price of the robot, they said that it "would have to be cheaper" \textit{(S01)}.

\subsubsection{Expected Use Cases (EC)} $ $

\textbf{(EC1; pos=11, neg=2)}: All participants in this study thought that the ZEST-E system would be useful for PWP. Several participants noted the system's potential to help individuals with mobility issues \textit{(S02)} and sedentary lifestyles \textit{(S04)}. Others thought that using ZEST-E would help PWP "maintain gains" made during physical therapy sessions \textit{(S01)} and would help "promote" and "restore" physical activity \textit{(S10)}. Some highlighted the potential of the system to help individuals with cognitive impairments \textit{(S02)}, but noted that "you would have to determine what level on the MoCA [cognitive examination] that they would need [to understand the robot's instructions]" \textit{(S11)}. One participant placed particular emphasis on the system's ability to adapt to individuals' needs:

\begin{quote}
    Because people with Parkinson’s, a lot of times they may have time frames during the day where they can go out and ride a bicycle. Then, in the afternoon, if they’re taking medications—those are timed and really crucial—they could have trouble getting up and down from a chair. Being able to still have those different levels of activities based off of what their current needs are, I think, would be helpful. \textit{(S01)}
\end{quote}

\textbf{(EC2; pos=11, neg=2)}: Additionally, all participants suggested that ZEST-E would be helpful for people with conditions other than PD. Suggestions were heavily focused on people with neurological conditions. Examples include, "any patient population—ortho, neuro" \textit{(S03)}, cardiology \textit{(S07)}, multiple sclerosis \textit{(S01)}, and stroke \textit{(S01, S08)}. In general, participants thought that robot-led exercise would be helpful for "anybody affected by any type of movement challenge" \textit{(S05)}, but others expressed concern that users "would have to be somebody who was relatively cognitively intact" \textit{(S03)}. Exercise specialists also expressed interest in using a system like ZEST-E as a gait assessment tool \textit{(S02, S03)}.

\textbf{(EC3; pos=7, neg=2)}: Many exercise specialists noted the potential for ZEST-E to help healthy people as well. One participant claimed that "pretty much anybody could benefit from those exercises" \textit{(S03)}, and another thought that "there are plenty of patients that could use it just for the accuracy of the feedback on the exercise" \textit{(S06)}. In the context of telemedicine, it was noted that ZEST-E could be useful for "harder to reach population, [such as] rural populations" \textit{(S04)}. Beyond adults, a specialist hypothesized that, "I think it could be good for kids, but for the safety of the robot, maybe not. Kids tend to be kind of destructive in my experience" \textit{(S02)}.

Some participants thought that ZEST-E's utility to healthy people might be limited. One participant said that ZEST-E would be useful for "more of a basic exercise, like just starting with warming up" \textit{(S08)}. Another expressed concern that, "as a healthy, independent person, there might not be a sincere need for me to use it... [unless] there was a very specific need within myself" \textit{(S07)}.

\textbf{(EC4; pos=11, neg=2)}: All 11 participants thought that the robot would be useful in a clinic or home environment. In the clinic, specialists focused on how the robot could reduce their workload. Example of integrating the robot into a clinical setting include using the robot for patient warm-ups \textit{(S08)} and being "more efficient with the clinic time" \textit{(S04)}. Others expressed interest in offloading low-level coaching to the robot: "[while exercising with ZEST-E], the exercise prescription is out of the therapist's hands and they're more focused on the skill [that the patient is working on]" \textit{(S10)} and "I could potentially be more productive, or doing something else at the same time but still being able to watch the qualitative nature of the exercise" \textit{(S06)}. Participants largely agreed that ZEST-E would fit naturally into a home health regimen. Several participants described the potential for ZEST-E to help people with exercises while they rest in bed \textit{(S03, S06)}. One person envisioned their use of ZEST-E as similar to existing home exercise protocols, but with better adherence reporting:

\begin{quote}
    I think it could be useful as a home exercise program if the physical therapist could choose what exercises specifically that it wanted to perform and then getting some data or feedback in the clinic or how the patient has been doing at home. It could give some compliance feedback as well to see, did they actually do the exercises, and how did they perform when doing them? \textit{(S11)}
\end{quote}

Yet, there was concern that "if somebody were to take it home at this stage, they wouldn't be able to operate it independently" \textit{(S03)}.

\textbf{(EC5; pos=5, neg=1)}: Participants identified several locations other than a clinic or at home where a robot similar to ZEST-E could be useful. Some examples include, "small group workouts with everybody that’s familiar with [the robot]" \textit{(S05)}, "rehab space, research spaces" \textit{(S07)}, "a skilled nursing facility or... in a community center type of senior center" \textit{(S01)}, and gymnasiums inside of "assisted livings and independent living facilities" \textit{(S01)}.







\section{DISCUSSION}


\subsection{Strengths of Robot-led Exercise}

ES identified many strengths and promising potentials for robot-led physical therapy, which fell under several broad themes that are examined in the following sections.

\subsubsection{Robots like ZEST-E have the potential to be useful in clinics, homes, and other spaces}

Several ES described use cases for ZEST-E inside the clinic. Benefits of including a robotic exercise aid in clinical treatment included offloading coaching warm-up exercises and providing low-level feedback on specific motions. Incorporating robotic exercise aids would also enable ES to work with more patients at the same time. Including a physical external target placed by the robot to "cue" stretching motions, as well as incorporating auditory feedback upon successfully reaching the target, encouraged participants to perform their full ROM. ES also hypothesized that exercising with ZEST-E could be helpful for others without PD, such as stroke survivors or even healthy older adults. 

ES highlighted many scenarios in which they could imagine using a system like ZEST-E outside of the clinic. Multiple ES thought that ZEST-E could be used for maintaining gains made in the clinic between clinic visits. ES were interested in deploying ZEST-E in shared health and living spaces, as well as in telemedicine contexts such as in their patients' homes and especially with rural populations. Also, using a robot to guide PT exercises creates an opportunity to collect and track rich, long-term health data inside and outside of the clinic, which can be used by ES to improve treatment plans. ES emphasized the ability of ZEST-E to work with people of varying degrees of mobility by leading seated as well as standing exercises, and they mentioned the possibility of including new exercises for individuals lying in bed. These results suggest that there are broad opportunities for robot-led PT inside and outside of the clinic.

\subsubsection{Robots like ZEST-E could provide effective and personalized care to PWP}

Participants reported that the system was adaptable to individuals, as shown by positive PIADS ratings (Figure \ref{fig:piads_tlx}) and high ratings for performance and therapeutic efficacy in the ZEQ (Figure \ref{fig:roq_zeq}). Further, ES expressed optimism that a system like ZEST-E would improve adherence to at-home exercise. Of specific interest was giving patients the ability to time their exercise sessions based on their on-off medication cycles throughout the day without having to leave their homes. These findings indicate that robots like ZEST-E could provide effective and personalized care to PWP.

\subsubsection{ZEST-E is engaging}

The multi-modal interaction between ZEST-E and participants resulted in ES rating the system as highly engaging. Sound effects, two-way verbal interactions, and video instructions were all cited as positive attributes of the system. ES highlighted the importance of encouragement delivered by the robot during exercises to boost engagement. Perceived enjoyment was rated highly (Figure \ref{fig:roq_zeq}), which indicates that the robotic exercise system was effective at engaging patients, and engagement ratings from the ZEST-E evaluative questionnaire (Figure \ref{fig:roq_zeq}) were also positive.

\subsection{Potential Improvements and Limitations}

A large focus of the ES during the semi-structured interviews was on suggestions for improving the ZEST-E system. We categorized their feedback into two broad groups that address distinct sets of limitations: quality of exercise instruction and ease of use.


\subsubsection{Improving the quality of robotic exercise coaching}

ES identified several features missing from ZEST-E that a human exercise coach would typically include. These enhancements could improve the ability of a robot exercise coach to guide sessions effectively, whether working under human supervision or operating independently.

First, ES expressed the need for ZEST-E to monitor participants' form during exercises and offer corrective feedback if form deviates from the ideal. Since each exercise was scored by ZEST-E based on the number of repetitions completed, users may be encouraged to cheat and not go through their full range of motion to score points more quickly. It was suggested that monitoring participants' form would also indicate high fatigue levels, so that ZEST-E could automatically stop the set when exercise form degraded due to fatigue. Also, monitoring fall risks was a concern among several ES, especially when considering deploying ZEST-E in environments outside of the clinic without someone to supervise present.  Incorporating a computer vision or wearable sensor-based exercise form feedback and/or fall risk detection system would make ZEST-E a more effective and safe exercise instructor.

Some ES thought that the current interaction scheme between ZEST-E and the user was "too robotic" or not interactive enough. All ZEST-E's spoken phrases were chosen by investigators before the study; while ZEST-E varied its selection of the spoken phrases, some users felt that ZEST-E was behaving in a way that felt inauthentic and pre-programmed, which reduced engagement. Also, ZEST-E was unable to handle requests for repeating instructions or clarifying instructions to address exercise form deviations. ZEST-E also could not make small talk during a session, which led to some participants evaluating the human-robot interaction as too "forced." Augmenting ZEST-E's conversation engine using a more sophisticated language processing scheme, such as a large language model (LLM), and an improved tone of voice could improve users' enjoyment and decrease confusion while interacting with ZEST-E.

While ES remarked favorably about the exercise designs, many expressed a desire for ZEST-E to lead a wider variety of exercises. Specific exercise suggestions included gait training, seated to standing transitions, overhead reaching, fine motor skills, or resistance training. However, many of these exercise types are not as well-suited to be led by a robot like ZEST-E. Because ZEST-E is lightweight and has a small footprint, exerting large forces on the robot would cause it to tip over \cite{kemp2022design}. This device is not suitable for resistance training, and it raises safety concerns for exercises such as gait and transfer training which involve a higher fall risk, since the robot cannot catch a falling participant. Additionally, ZEST-E's reachable workspace is between the floor and approximately 1.3m above the ground \cite{kemp2022design}, which limits the height at which ZEST-E can place the target for overhead reaching exercises. The development of mobile robots suitable for gait training warrants further investigation.

\subsubsection{Improving ease of use}

ES identified many ways to improve the ease of use while operating ZEST-E. From the perspective of an ES using ZEST-E in a clinic, participants highlighted the need to minimize robot setup time and maximize ease of operation. This includes minimizing the number of inputs to the system that the operator must make, as well as increasing the robot's level of autonomy from leading single exercise sets to leading entire routines. In addition, while ZEST-E is able to be transported by a single human without peripheral equipment \cite{kemp2022design}, ES expressed a desire for an even more portable robot. Despite these suggestions, the system in its current state had high ratings for its Perceived Ease of Use in the ROQ (Figure \ref{fig:roq_zeq}).

Several other key improvements were also identified to make the system easier for PWP to use on their own outside of the clinic. In this context, ES provided further emphasis that ZEST-E should operate mostly autonomously, especially for older adults who have lower levels of technological literacy. Alternative modalities to interact with the robot, such as a direct voice interface to launch exercises, would reduce barriers for PWP using a system like ZEST-E independently. Including further accessibility options, such as adding subtitles to ZEST-E's screen for spoken instructions and adding other languages, would further decrease barriers for adoption. These suggestions were supported by mixed ratings for ease of use for patients in the ZEST-E Evaluative Questionnaire and in the broad spread of Perceived Usefulness ratings from the ROQ (Figure \ref{fig:roq_zeq}).

\section{CONCLUSION}

Integrating therapeutic robotic systems into established rehabilitative programs for PWP shows significant promise. Our mixed-methods study indicates that a platform such as ZEST-E can deliver engaging, personalized support in clinics, homes, and other environments. Yet, further refinement is needed to strengthen the robot’s instructional abilities and enable smooth integration into routine therapy. Future research, particularly on therapist–patient–robot triad interactions in a clinical setting, could reveal additional opportunities to improve robotic therapy for PWP.

\section*{FUNDING}

We thank the Toyota Research Institute (grant GR00010714) and the McCamish Parkinson's Disease Innovation Program (grant DE00021631) for funding this research. 

\section*{ACKNOWLEDGEMENTS}

We thank Isabel Scheib and David Denton for their feedback and contributions to this project.

\section*{Competing Interests}

Charles C. Kemp was an associate professor at the Georgia Institute of Technology when contributing to this research. He now works full-time for Hello Robot Inc., which sells the Stretch RE1.



\bibliographystyle{ieeetr} 
\bibliography{ref}  

\end{document}